\documentclass[runningheads]{llncs}
\usepackage{graphicx}
\usepackage{amsmath,amssymb} % define this before the line numbering.
\usepackage{color}
\usepackage{multirow}
\usepackage{pifont}% http://ctan.org/pkg/pifont
\newcommand{\cmark}{\ding{51}}%
\newcommand{\xmark}{\ding{55}}%
\usepackage{bm}
\begin{document}
\title{Person Search by Multi-Scale Matching} 
% Replace with your title

% Replace with a meaningful short version of your title
%
\author{Xu Lan\inst{1} %\orcidID{0000-0001-5927-0986} 
	\and
	Xiatian Zhu\inst{2} % \orcidID{0000-0002-9284-2955} 
	\and
	Shaogang Gong\inst{1} %\orcidID{0000-0001-8156-2299}
}
%
%Please write out author names in full in the paper, i.e. full given and family names. 
%If any authors have names that can be parsed into FirstName LastName in multiple ways, please include the correct parsing, in a comment to the volume editors:
%\index{Lastnames, Firstnames}
%(Do not uncomment it, because you may introduce extra index items if you do that, we will use scripts for introducing index entries...)
\authorrunning{A. Author and B. Author}
% Replace with shorter version of the author list. If there are more authors than fits a line, please use A. Author et al.
%

\institute{$^1$ Queen Mary University of London,\\
	\email{x.lan@qmul.ac.uk}, \email{s.gong@qmul.ac.uk}	\\
	$^2$ Vision Semantics Ltd \\
	\email{eddy@visionsemantics.com}
	%	\OrcidID{0000-0002-9284-2955}
	%, \url{https://orcid.org/0000-0002-9284-2955}
}

\maketitle              % typeset the header of the contribution
\begin{abstract}
We consider the problem of person search in unconstrained scene images.
Existing methods usually focus on improving the person 
detection accuracy to mitigate 
negative effects imposed by misalignment, mis-detections, and false alarms
resulted from noisy people auto-detection. 
In contrast to previous studies, we show that sufficiently reliable person instance cropping 
is achievable by slightly improved state-of-the-art deep learning object detectors (e.g. Faster-RCNN), and the under-studied multi-scale matching problem in {person search} is % potentially 
a more severe barrier.
In this work, we address this multi-scale person search challenge by proposing a 
{\em Cross-Level Semantic Alignment} (CLSA) {deep learning approach
	capable of learning} more discriminative identity feature representations
in a unified end-to-end model.
This is realised by exploiting the 
in-network feature pyramid structure of a deep neural network
enhanced by a novel cross pyramid-level semantic alignment 
loss function. 
This favourably eliminates the need 
for constructing a computationally expensive image pyramid
and a complex multi-branch network architecture.
Extensive experiments 
show the modelling advantages and performance superiority of CLSA %model
over the state-of-the-art person search and multi-scale matching
methods on two % publicly available 
large person search benchmarking datasets: CUHK-SYSU and PRW.
%\dots
\keywords{Person Search; Person Detection and Re-Identification; Multi-Scale Maching; 
	Feature Pyramid; Image Pyramid; Semantic Alignment.}
\end{abstract}
\section{Introduction}

\begin{figure}
	\centering
	\includegraphics[height=5.5cm]{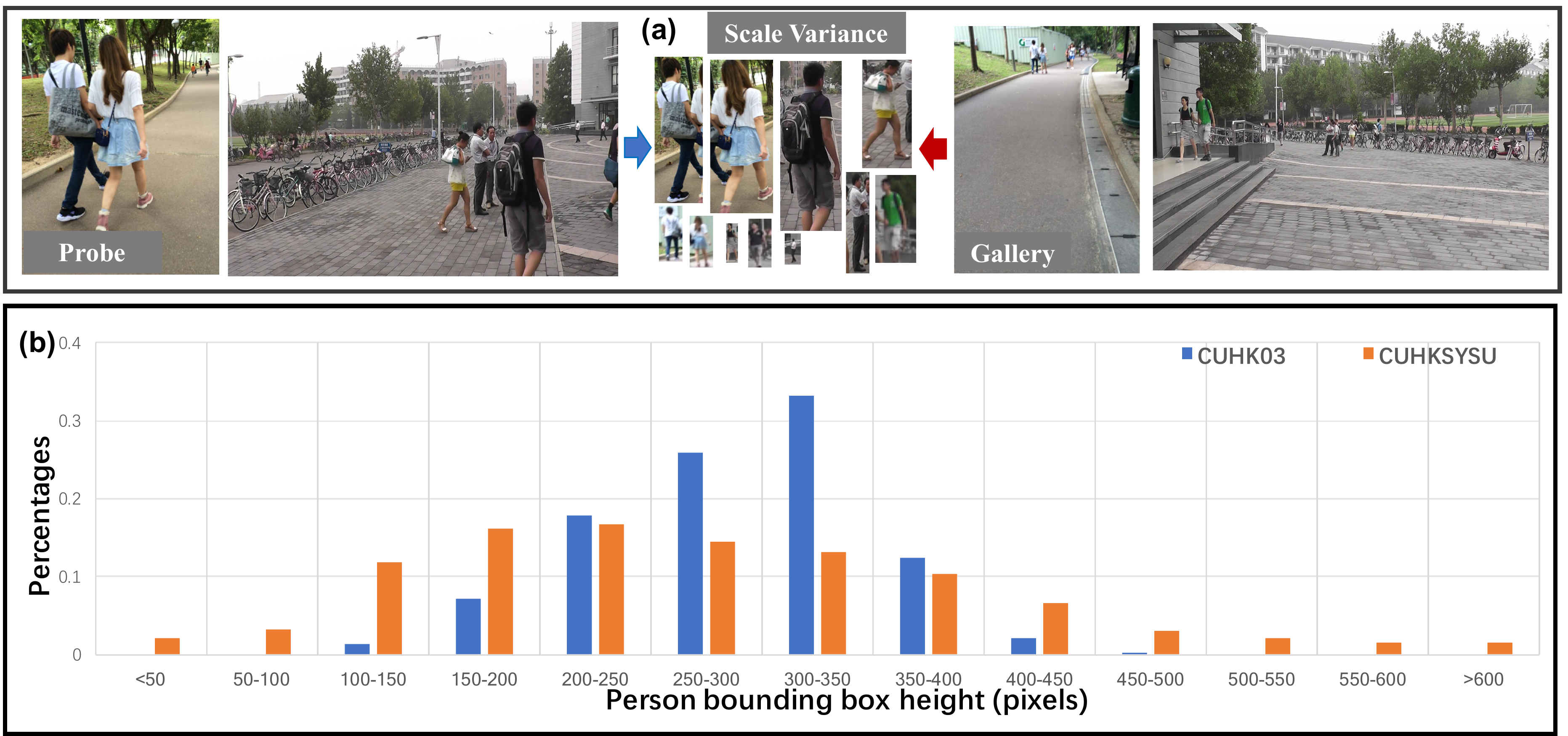}
	\caption{Illustration of 
		the intrinsic multi-scale matching challenge 
		in person search.
		(a) Auto-detected person bounding boxes vary significantly 
		in scale.
		(b) The person scale distribution of CUHK-SYSU (person search benchmark) 
		covers a much wider range than manually refined CUHK-03 (person re-id benchmark).
	}
	\label{fig:scale_change}	
\end{figure}

Person search aims to find a probe person
in a gallery of whole unconstrained scene images \cite{xiao2017joint}. 
It is an extended form of person re-identification (re-id)
\cite{gong2014person} {by} additionally considering
the requirement of %not only match the identity class, but also
automatically detecting people in the scene images
besides matching the identity classes.
Unlike the conventional person re-id problem assuming the gallery images
as either manually cropped or carefully filtered auto-detected bounding boxes
\cite{xiao2016learning,li2017person,chen2017beyond,hermans2017defense,lan2017deep,zhang2017deep,li2018harmonious,wang2014person,wang2018person,chang2018multi,zhang2017deep}, % most with similar spatial scales (resolution),
person search deals with raw unrefined detections % are not specially refined 
with
% contain 
many false cropping and unknown degrees of
misalignment. This yields a more challenging matching problem especially in the process of
person re-id. 
Moreover, 
auto-detected person boxes often vary more significantly in scale (resolution)
than the conventional person re-id benchmarks (Fig. \ref{fig:scale_change}(b)),
due to the inherent uncontrolled distances between persons and cameras
(Fig. \ref{fig:scale_change}(a)).
It is therefore intrinsically a {\em multi-scale matching} problem.
However, this problem is currently under-studied in person search
\cite{xiao2017joint,zheng2016person,liu2017neural}.

In this work, we aim to address the multi-scale matching problem in person
search.
We show that this is a significant factor in improving the model matching performance, given the
arbitrary and unknown size changes of persons in auto-detected bounding boxes.
However, existing methods \cite{xiao2017joint,zheng2016person,liu2017neural}
focus on the person detection and localisation in scene images,
which
turns out not to be a severe bottleneck for the overall search
performance as indicated in our experiments. 
For example, using the 
ground-truth person bounding boxes only brings a Rank-1 gain of
1.5\% alongside employing ResNet-50 \cite{he2016deep} for person search on the
CUHK-SYSU benchmark \cite{xiao2017joint}.
In contrast, with the same ResNet-50 model, our proposed multi-scale matching learning 
improves the person search Rank-1 rate by 6.0\% on the
same benchmark (Fig.~\ref{fig:mAP-decreased}).

We make three {\bf contributions} in this study:
{\bf (1)}
We identify the multi-scale matching problem in person search --
an element missing in the literature but found to be significant 
for improving the model performance.
{\bf (2)}
We formulate a {\em Cross-Level Semantic Alignment} (CLSA) deep learning
approach to addressing the multi-scale matching challenge.
This is based on learning an
end-to-end in-network feature pyramid representation 
with superior robustness in coping with
variable scales of {auto-detected person bounding boxes.}
{\bf (3)}
We improve the Faster-RCNN model
for more reliable person localisation in uncontrolled scenes,
facilitating the overall search performance.
Extensive experiments on two benchmarks 
CUHK-SYSU \cite{xiao2017joint} and PRW \cite{zheng2016person}
show the person search advantages of the proposed CLSA
over state-of-the-art methods,
improving the best competitor by 
{7.3\%} on CUHK-SYSU
and {11.9}\% on PRW in Rank-1 accuracy.

\section{Related Work}

\noindent{\bf Person Search }
Person search is a recently introduced problem
%The task is to 
of matching a probe person bounding box against 
a set of gallery whole scene images
\cite{xiao2017joint,zheng2016person}.
This is challenging due to 
the uncontrolled false alarms, mis-detections,
and misalignment emerging in the auto-detection process.
In the literature, there are only a handful of person search works
\cite{xiao2017joint,zheng2016person,liu2017neural}.
Xiao et al. \cite{xiao2017joint}
propose a joint detection and re-id 
deep learning model for seeking their complementary 
benefits.
Zheng et al. \cite{zheng2016person}
study the effect of person detection 
on the identity matching performance.
%Recently, 
Liu et al. \cite{liu2017neural}
consider recursively search refinement 
to more accurately locate the target person
in the scene.
While existing methods focus on detection enhancement,
%in the search process,
we show that 
by a state-of-the-art deep learning object detector with small improvements,
person localisation is not a big limitation.
Instead, the multi-scale matching problem 
turns out a more severe challenge in person search. % (Fig. \ref{fig:scale_change}).
In other words, solving the multi-scale problem
is likely to bring more performance gain
than improving person detection (Fig. \ref{fig:mAP-decreased}(c)).

\noindent{\bf Person Re-Identification }
Person search is essentially an extension of 
the conventional person re-id problem \cite{gong2014person}
with an additional requirement of automatic person detection
in the scenes.
Given the manual construction nature of re-id datasets,
the scale diversity of gallery images tends to be
restricted.
It is simply harder for humans to verify and label the person identity of small bounding boxes,
therefore leading to the selection and labelling bias towards large boxes (Fig. \ref{fig:scale_change}(b)).
Consequently, the intrinsic {multi-scale matching} challenge is {\em artificially} suppressed in
re-id benchmarks,
hence losing the opportunity
to test the real-world model robustness.
Existing re-id methods can mostly afford to ignore the problem of 
multi-scale person bounding boxes in algorithm design.
Whilst extensive efforts have be made
to solving the re-id problem
\cite{wang2014person,li2014deepreid,cheng2016person,wang2016human,xiao2016learning,li2017person,chen2017beyond,zheng2015scalable,lan2017deep,zhu2017fast,wang2018person,jiao2018deep,li2018harmonious,li2018Unsupervised,wang2018transferable,chen2017person,chen2018person},
there are only limited works
considering multi-scale matching
\cite{chen2017person,liu2016multi}.
%
%Compared to these existing multi-scale re-id learning models,
Beyond all these existing methods,
our CLSA is designed specially to explore the 
in-network feature pyramid in deep learning
for more effectively 
solving the under-studied 
multi-scale challenge in person search.

\section{Cross-Level Semantic Alignment for Person Search}

\begin{figure}
	\centering
	\includegraphics[height=3.8cm]{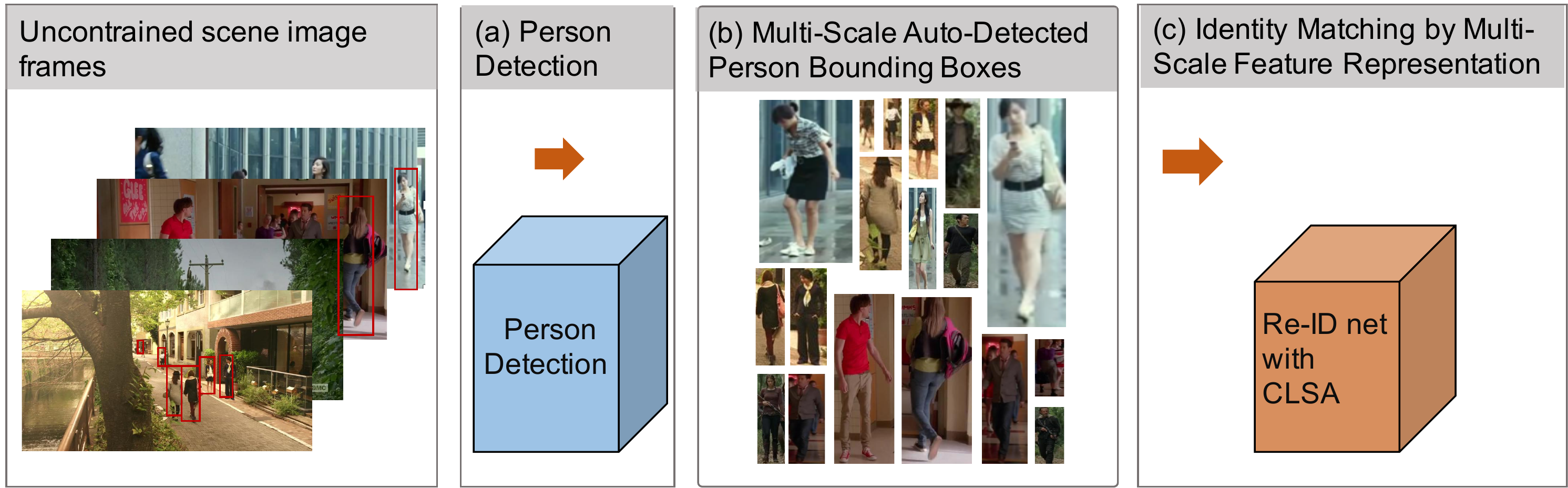}
	
	\caption{Overview of the proposed multi-scale learning person search framework. 
		(a) Person detection for cropping people
		from the whole scene images
		at (b) varying scales (resolutions).
		(c) Person identity matching is then conducted by a re-id model.
	}
	\label{fig:person_search}	
\end{figure}

We want to establish a person search system % {\color{red}to be}
capable of automatically detecting 
and matching persons
in unconstrained scenes with any probe person.
With the arbitrary distances between people and cameras in public space,
person images are inherently captured at varying scales and resolutions.
This raises the multi-scale matching challenge.
To overcome this problem, we formulate a 
Cross-Level Semantic Alignment (CLSA) deep learning approach.
An overview of the CLSA is illustrated in Fig. \ref{fig:person_search}.
The CLSA contains two components:
(1) Person detection which 
locates all person instances in the gallery scene images
for facilitating the subsequent identity matching.
(2) Person re-identification which 
matches the probe image against a large number of 
arbitrary scale gallery person bounding boxes 
(the key component of CLSA).
We provide the component details below.

\subsection{Person Detection}
\label{sec:det}
As a pre-processing step, person detection is important 
in order to achieve accurate search \cite{xiao2017joint,zheng2016person}.
We adopt the Faster-RCNN model \cite{ren2015faster} 
as the CLSA detection component, due to its strong capability 
of detecting varying sized objects in unconstrained scenes. 
%for reducing the mis-detection rate.
%
%
To further enhance person detection performance and efficiency, 
{we introduce a number of design improvement on the original model.}
%we make a number of model optimisation refinements
%on the original Faster-RCNN model. 
%
{\bf (1)} %{\color{red} Following \cite{chen17implementation}}, 
Instead of using the conventional RoI (Region of Interest) pooling layer, 
we crop and resize the region feature maps to $14\!\times\!14$ in pixel, 
and further max-pool them to $7\!\times\!7$ 
for gaining better efficiency \cite{chen17implementation}.  
% \cite{pytorchfasterrcnn,chen2017implementation}.
% \cite{pytorchfasterrcnn,chen2017implementation}. 
% 
{\bf(2)} After pre-training the backbone ResNet-50 net on ImageNet-1K, 
we fix the 1$^\text{st}$ building-block (the 1$^\text{st}$ 4 layers) 
in fine-tuning on the target person search data. 
This allows to preserve the shared low-level features learned from
larger sized source data
whilst simultaneously adapting the model to target data.
%which is same as  \cite{pytorchfasterrcnn}.
{\bf(3)} 
We keep and exploit all sized proposals for reducing the mis-detection
rate at extreme scales in uncontrolled scenes {before the Non-Maximum Suppression (NMS) operation}. % due to it will help detect small person and increasing our recall rate.
In deployment, 
we consider all detection boxes
scored above $0.5$, % unlike \cite{zheng2016person} 
rather than extracting a fixed number of boxes
from each scene image \cite{zheng2016person}.
This is because the gallery scene images may contain varying (unknown in priori) number of people.

\subsection{Multi-Scale Matching by Cross-Level Semantic Alignment}
\label{sec:CLSA}
Given auto-detected person bounding boxes at arbitrary scales from the gallery scene images,
we aim to build a person identity search model
robust for multi-scale matching.
To this end, we explore the seminal image/feature pyramid concept \cite{adelson1984pyramid,lazebnik2006beyond,lowe2004distinctive,dalal2005histograms}.
Our motivation is that
a single-scale feature
representation blurs salient and discriminative information at different scales useful in
person identity matching;
And a pyramid representation allows to be
``scale-invariant'' (more ``scale insensitive'')
in the sense that a scale change in matching images is counteracted by a scale shift 
within the feature pyramid.

\noindent{\bf Build-In Feature Pyramid }
We investigate the multi-scale feature representation learning 
in deep Convolutional Neural Network (CNN) to exploit
the built-in feature pyramid structure formed on a single input
image scale.
Although CNN features have shown to be more robust to variance in image scale,
pyramids are still effective in seeking more accurate detection and recognition results \cite{lin2017feature}.

\begin{figure}
	\centering
	\includegraphics[width=1\linewidth]{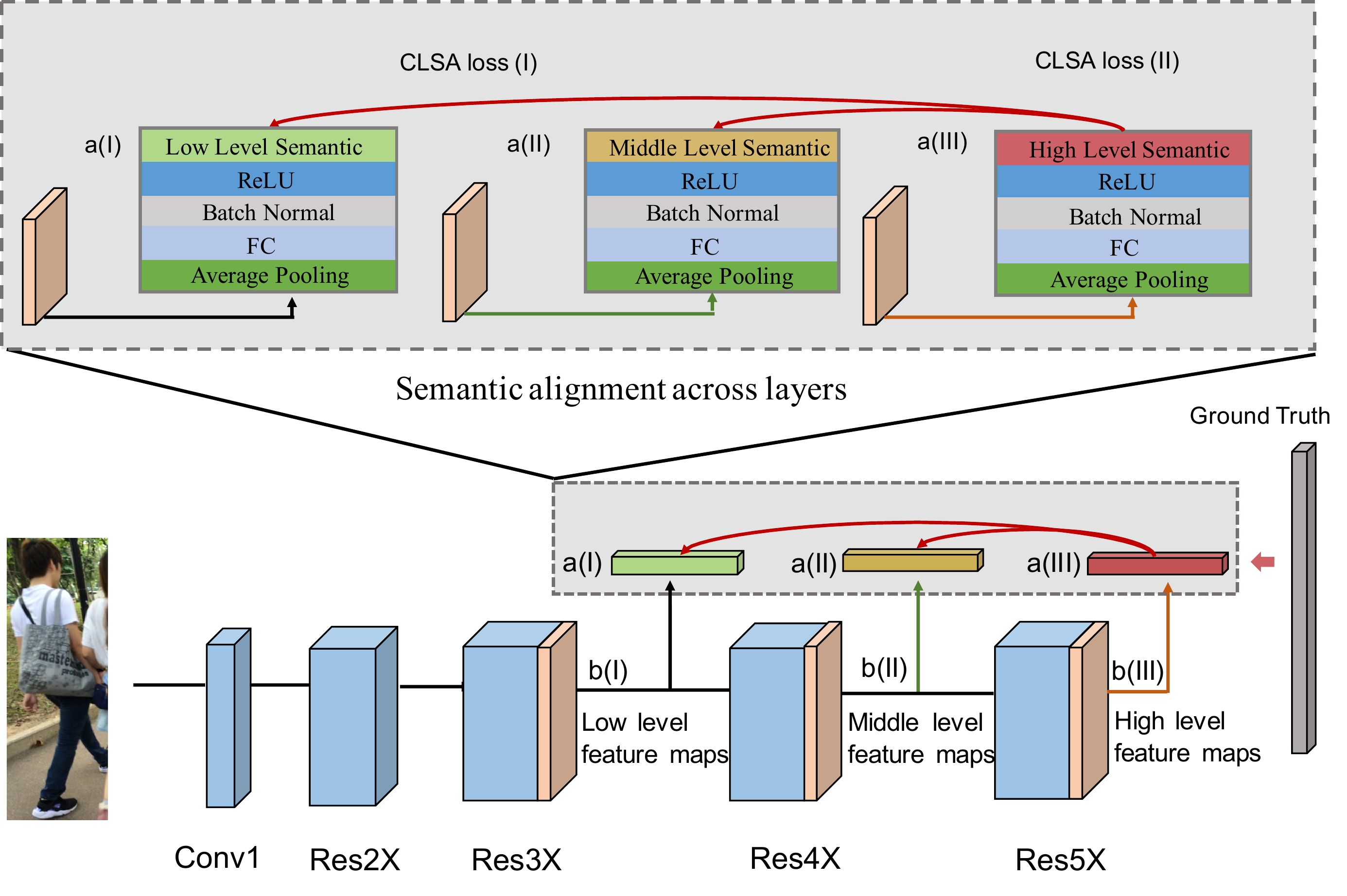}
	\caption{
		Overview of the proposed Cross-Level Semantic Alignment (CLSA) approach
		in a ResNet-50 based implementation.
	}
	\label{fig:scale-reid}	
\end{figure}

For the CNN architecture,
we adopt the state-of-the-art ResNet-50 \cite{he2016deep} 
as the backbone network (Fig. \ref{fig:scale-reid})
of the identity matching component.
%due to its capacity for deeper/stronger network design 
%whilst retaining a smaller model parameter size.
%For training this matching model, 
In this study, we particularly leverage the feature pyramid hierarchy
with low-to-high levels of semantics from bottom to top layers,
%
%Such a feature pyramid is 
automatically established in model learning optimisation
\cite{zeiler2014visualizing}.
Given the block-wise net structure in ResNet-50,
we build a computationally efficient $K$-levels feature pyramid
using the last conv layer of top-$K$ ($K\!=\!3$ in our experiments) blocks.
The deepest layer of each block is supposed to have the most semantic features.

Nonetheless, it is not straightforward to exploit the ResNet-50 feature hierarchy.
This is because the build-in pyramid has large semantic gaps 
across levels due to the distinct depths of layers.
The features from lower layers are less discriminative
for person matching therefore likely hurt the overall representational capacity
if applied jointly with those from higher layers.

\noindent{\bf Cross-Level Semantic Alignment }
To address the aforementioned {problems}, 
we improve the in-network feature pyramid 
by introducing a Cross-Level Semantic Alignment (CLSA) learning mechanism.
The aim is to achieve a feature pyramid with all levels encoding the desired high-level 
person identity semantics.
Formally, to train our person identity matching model,
we adopt the softmax Cross-Entropy (CE) loss function
to optimise an identity classification task.
The CE loss on a training person bounding box $(\bm{I}, y)$ is computed as: 
\begin{equation}
\mathcal{L}_\text{ce}=
{-}{\log}\Big(\frac{\exp({\bm{W}_{y}^{\top} {\bm x}})}
{\sum_{i=1}^{|\mathcal{Y}|}\exp({\bm{W}_{i}^{\top} {\bm x}})}\Big)
\label{eq:CE_loss}
\end{equation}
where $\bm{x}$ specifies the feature vector of $\bm{I}$
by the last layer,
$\mathcal{Y}$ the training identity class space,
and % i.e. ${\bm x}_i=f(X_i; \theta_{F})$
$\bm{W}_{y}$ the $y$-th ($y\in \mathcal{Y}$) class prediction function parameters. 

In our case, 
$\bm{x}$ is the top pyramid level, also denoted as $\bm{x}^K$.
For anyone of the top-$K$ ResNet blocks, we 
obtain $\bm{x}$ by applying an average pooling layer and a FC layer
on the output feature maps (Fig. \ref{fig:scale-reid} (b)).
Consider the different feature scale distributions across layers \cite{liu2015parsenet},
we further normalise $\bm{x}$
by batch normalisation and ReLU non-linearity. 
In this way, we compute the feature representations for all $K$ pyramid layers
$\{\bm{x}^1, \cdots,\bm{x}^K\}$.

Recall that we aim to render all levels of feature representations identity semantic.
To this end, we first project each of these features $\{\bm{x}^1, \cdots,\bm{x}^K\}$ 
by a FC layer into the 
identity semantic space with the same dimension as $\mathcal{Y}$.
The resulted semantic class probability vectors are denoted as 
$\{\bm{p}^1, \cdots,\bm{p}^K\}$
with $\bm{p}^k \!=\! [p_1^k, \cdots, p_{|\mathcal{Y}|}^k]$,
$k\in\{1,\cdots,K\}$.
To transfer the strongest semantics from the
top ($K$-th) pyramid level to a lower ($s$-th) level,
we introduce a Kullback-Leibler divergence based Cross-Level Semantic Alignment (CLSA) loss formulation
inspired by knowledge distillation \cite{hinton2015distilling}:
\begin{equation} 
\label{eq:kl_loss}
\mathcal{L}_\text{clsa}(s)= \sum_{j=1}^{|\mathcal{Y}|}   {\tilde{p}_j^K} \log \frac {\tilde{p}_j^K}{{\tilde{p}_j^s}}.
\end{equation}
where $\tilde{p}_j^k$ is 
a {\em softened} per-class prediction semantic score
obtained by
\begin{equation}
\label{eq:softmax_soft}
\tilde{p}_j^k = \frac{\exp(p_j^k/T)} {\sum_{j=1}^{|\mathcal{Y}|} \exp(p_j^k/T)},
\end{equation}
where the temperature parameter $T$ controls the
softening degree (higher values meaning more softened predictions). 
{We set T=3 %in our experiments 
	following the suggestion in \cite{hinton2015distilling}}.
To enable end-to-end deep learning, we add this CLSA loss 
on top of the conventional CE loss (Eq \eqref{eq:CE_loss}):
\begin{equation} 
\label{eq:Total_loss}
\mathcal{L}= \mathcal{L}_\text{ce} + T^2 \sum_{s=1}^{K-1}\mathcal{L}_\text{clsa}(s)
\end{equation}
where $T^2$ serves as a weighting parameter between the two loss terms. %  in Eq \eqref{eq:softmax_soft}

\noindent{\bf Identity Matching by CLSA Feature Pyramid }
{%\color{red} 
	In deployment, we first compute a CLSA feature pyramid
	by forward propagating any given person bounding box image.
	We then concatenate the feature vectors of all pyramid levels as the final representation for person re-id matching.}

\noindent\textbf{\em Remarks }
The CLSA is similar in spirit to
a few person re-id matching methods
\cite{chen2017person,liu2016multi}.
However, these methods adopt the image pyramid scheme, in contrast to the CLSA leveraging
the in-network feature pyramid on a single image scale 
therefore more efficient.
The FPN model \cite{lin2017feature}
also exploits the build-in pyramid.
The CLSA differs from FPN in a number of fundamental ways:
(1) FPN focuses on object detection and segmentation,
whilst CLSA aims to address fine-grained identity recognition and matching.
(2) FPN additionally performs feature map unsampling 
hence less efficient than CLSA. % without this need.
(3) CLSA performs semantic alignment and transfer in the 
low-dimensional class space,
in comparison to more expensive FPN's feature alignment.
We will evaluate and compare these multi-scale learning methods
against CLSA in our experiments (Table \ref{table:scale_method}).

\section{Experiments}

\begin{figure}
	\centering
	\includegraphics[width=1.0\linewidth]{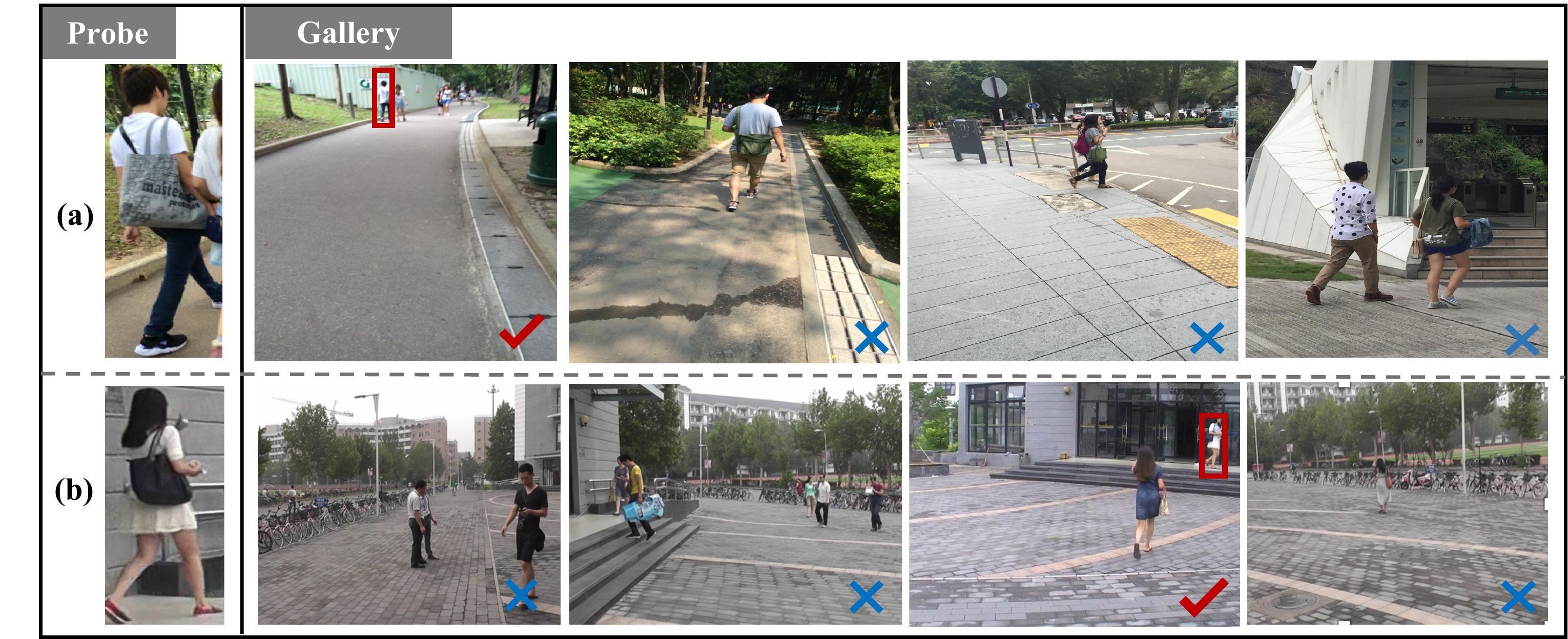}
	%\vskip -0.2cm
	\caption{
		Example probe person and unconstrained scene images 
		on (a) CUHK-SYSU \cite{xiao2017joint} and (b) PRW \cite{zheng2016person}.
		Green bounding box: the ground truth probe person in the scene.
		\cmark: Contain the probe person.
		\xmark: Not contain the probe person.
		%		It is typical that people appearing in unconstrained scenes
		%		were captured at arbitrary scales by surveillance camera views.
	}
	\label{figure:dataset_ex}	
\end{figure}

\noindent{\bf Datasets }
To evaluate the CLSA, we selected two person search benchmarks: 
CUHK-SYSU \cite{xiao2017joint} and PRW \cite{zheng2016person}.
We {adopted} the standard evaluation setting as summarised in Table \ref{table:data_split}.
In particular, the CUHK-SYSU dataset
contains 18,184 scene images, 8,432 labelled person IDs, and 96,143 annotated person bounding boxes. 
Each probe person appears in two or more scene gallery images captured from different 
locations. 
The training set has 11,206 images and 5,532 probe persons. 
Within the testing set, the probe set includes 2,900 person bounding boxes 
and the gallery contains a total of 6,978 whole scene images.
The PRW dataset provides a total of  11,816 video frames and 43,110 person bounding boxes. 
The training set 
has 482 different IDs from 5,704 frames.  
The testing set contains 2,057 probe people along with a gallery of 6,112 scene images.
In terms of bounding box scale, 
CUHK-SYSU and PRW range from $37\!\times\!13$ to $793\!\times\!297$, 
and $58\!\times\!21$ to $777\!\times\!574$, respectively. 
This shows the two person search datasets present the intrinsic multi-scale challenge.
Example images are shown in Fig. \ref{figure:dataset_ex}.

\noindent{\bf Performance Metrics }
For person detection, a person box is considered as correct 
if overlapping with the ground truth over 50\% \cite{xiao2017joint,zheng2016person}.
For person identity matching or re-id, 
we adopted the {Cumulative Matching Characteristic} (CMC)
and mean Average Precision (mAP). % to measure re-id test accuracy.
The CMC is computed on each individual rank $k$ as the probe cumulative percentage of truth matches appearing at ranks $\leq$$k$.
The mAP measures the recall of multiple truth matches,
computed by first computing the area under the
Precision-Recall curve for each probe, then calculating the mean of
Average Precision over all probes \cite{zheng2015scalable}.

\begin{table}[h]
	\centering
	\setlength{\tabcolsep}{0.08cm}
	\caption{Evaluation setting, data statistics, and person bounding box scale of 
		the CUHK-SYSU and PRW benchmarks.
		Bbox: Bounding box. 
	}
	\label{table:data_split}
	%\vskip -0.2cm
	%	\tabcolsep=0.1cm
	\begin{tabular}{l||c|c|c|c|c|c|c|c}
		\hline 
		\multirow{2}{*}{Dataset} 
		& \multirow{2}{*}{Images}
		&	\multirow{2}{*}{Bboxes}& \multirow{2}{*}{IDs} &\multirow{2}{*}{\bf Bbox Scale} & \multicolumn{2}{c|}{ID Split}&\multicolumn{2}{c}{Bbox Split} \\
		\cline{6-9}
		& & &&&Train & Test & Train & Test \\
		\hline \hline
		CUHK-SYSU
		& 18,184
		& 96,143 & 8,432 &$37\!\times\!13$$\sim$$793\!\times\!297$& 5,532&2,900&55,272&40,871\\
		\hline
		PRW
		& 11,816
		& 43,110& 932 & $58\!\times\!21$$\sim$$777\!\times\!574$&482&450& 18,048&25,062 \\
		\hline
	\end{tabular}
	%\vspace{-0.2cm}
\end{table}

\noindent{\bf Implementation Details }
We adopted the Pytorch framework \cite{paszke2017pytorch} {to} conduct 
all the following experiments. For training the person detector component, 
we adopted the SGD algorithm with the momentum set to 0.9, 
the weight decay to 0.0001，
the iteration to 110,000,
and the batch size to 256. 
We initialised the learning rate at 0.001, 
with a decay factor of 10 at every 30,000 iterations. 
%We trained {\color{red}it} for 110,000 iterations at a batch size of 256. 
%
For training the identity matching component, 
{we used both annotated and detected (over $50\%$ Intersection over Union (IoU) with the annotated and sharing the identity labels) boxes %as the training set
	as \cite{zheng2016person}.}
We set the momentum to 0.9, 
the weight decay to 0.00001,
the batch size to 64,
and the epoch to 100.
The initial learning rate was set at 0.01, 
and decayed by 10 at every 40 epochs.
All person bounding boxes were resized to $256 \times 128$ pixels.
To construct the in-network feature pyramid,
we utilised the top 3 (Res3x, Res4x, Res5x) blocks
in our final model implementation, i.e. $K\!=\!3$ in Eq. \eqref{eq:Total_loss}.
We also evaluated other pyramid constructing ways in the component analysis
(Sec. 4.3).

\subsection{Comparisons to State-Of-The-Art Person Search Methods}		

We compared the proposed CLSA method with two groups of existing person search approaches:
(1) Three most recent state-of-the-art methods
(NPSM \cite{liu2017neural}, OIM \cite{xiao2017joint}, CWS \cite{zheng2016person}); and
(2) Five popular person detectors 
(DPM \cite{felzenszwalb2010object}, 
ACF \cite{dollar2014fast}, 
CCF \cite{yang2015convolutional}, 
LDCF \cite{nam2014local}, and R-CNN \cite{girshick2014rich}) 
with hand-crafted
(BoW \cite{zheng2015scalable}, 
LOMO \cite{liao2015person}, 
DenseSIFT-ColorHist (DSIFT) \cite{zhao2013unsupervised}) 
or deep learning (IDNet \cite{xiao2017joint})
features based 
re-id metric learning methods (KISSME \cite{koestinger2012large}, XQDA \cite{liao2015person}).

\begin{figure}
	\centering
	\includegraphics[width=0.99\linewidth]{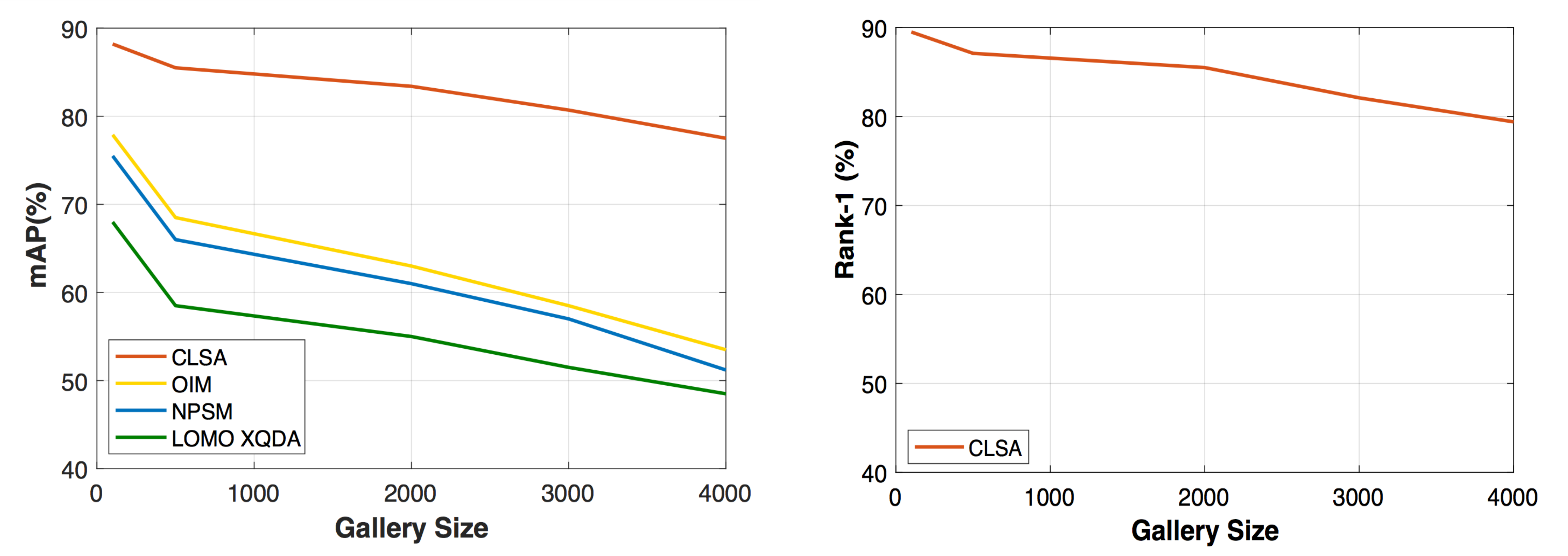}
	%	\vskip -0.2cm
	\caption{Model scalability evaluation
		over different gallery search sizes
		on CUHK-SYSU.
	}
	\label{figure:gallery_size}	
\end{figure}

\noindent{\bf Evaluation on CUHK-SYSU } 
Table \ref{table:cuhk_eval} reports the person search performance on CUHK-SYSU with 
the standard gallery size of 100 scene images. 
It is clear that the CLSA significantly outperforms all other competitors.
For instance, the CLSA surpasses the top-2 alternative models NPSM and OIM (both are end-to-end deep learning models) by 
7.3\% (88.5-81.2) and 
9.8\% (88.5-78.7) in Rank-1, 
9.3\% (87.2-77.9) and 
11.7\% (87.2-75.5) in mAP, respectively.
The performance margin {of CLSA against} other non-deep-learning methods
is even larger, due to that these models rely on less discriminative
hand-crafted features without the modelling advantage of 
jointly learning stronger representation and matching metric model.
{This shows the overall performance superiority of the CLSA over
	current state-of-the-art methods, 
	thanks to the joint contributions of improved person detection} model (see more details below)
and the proposed multi-scale deep feature representation learning mechanism.

{%\color{red}
	To evaluate the model efficiency, we conducted a person search test
	among 100 gallery images on CUHK-SYSU.
	We deployed a desktop with a Nvidia Titan X GPU.
	Applying CLSA, OIM, and NPSM takes 1.2, 0.8, and 120 seconds, respectively.
	This indicates that the performance advantages of our CLSA
	do not sacrifice the model efficiency.}

\begin{table}[h]
	\centering
	\setlength{\tabcolsep}{0.6cm}
	\caption{Evaluation on CUHK-SYSU. Gallery size: 100 scene images.
		The best and second-best results are in red and blue.}
	\label{table:cuhk_eval}

	\begin{tabular}{|c|c|c|}
		\hline %\noalign{\smallskip}
		Method &Rank-1 (\%)& mAP (\%) \\
		\hline \hline
		ACF\cite{dollar2014fast}+DSIFT\cite{zhao2013unsupervised}+Euclidean&25.9&21.7\\ 
		ACF\cite{dollar2014fast}+DSIFT\cite{zhao2013unsupervised}+KISSME\cite{koestinger2012large}&38.1&32.3\\
		%ACF+Bow\cite{zheng2015scalable}+Cosing&48.4&42.4  \\
		ACF\cite{dollar2014fast}+LOMO\cite{liao2015person}+XQDA\cite{liao2015person}& 63.1&55.5 \\		
		\hline
		CCF\cite{yang2015convolutional} +DSIFT\cite{zhao2013unsupervised}+Euclidean &11.7&11.3 \\
		CCF\cite{yang2015convolutional}+DSIFT\cite{zhao2013unsupervised}+KISSME\cite{koestinger2012large}   &13.9&13.4 \\
		%CCF+BoW+Cosine      &29.3&26.9 \\
		CCF\cite{yang2015convolutional}+LOMO\cite{liao2015person}+XQDA\cite{liao2015person}     &46.4&41.2 \\
		CCF\cite{yang2015convolutional}+IDNet\cite{xiao2017joint} &57.1&50.9 \\
		\hline
		CNN\cite{ren2015faster}+DSIFT\cite{zhao2013unsupervised}+Euclidean &39.4&34.5\\
		CNN\cite{ren2015faster}+DSIFT\cite{zhao2013unsupervised}+KISSME\cite{koestinger2012large}   & 53.6&47.8\\
		%CNN+BoW+Cosine   & 62.3 &56.9 \\
		CNN\cite{ren2015faster}+LOMO\cite{liao2015person}+XQDA\cite{liao2015person}  &74.1 &68.9 \\
		CNN\cite{ren2015faster}+IDNet\cite{xiao2017joint}         &74.8 &68.6\\
		OIM\cite{xiao2017joint}&78.7&75.5 \\
		NPSM\cite{liu2017neural}&\bf\color{blue}81.2&\bf\color{blue}77.9\\
		\hline
		\bf CLSA &{\bf\color{red}88.5}&{\bf\color{red}87.2}\\
		\hline
	\end{tabular}
\end{table}

To test the model performance scalability,
we further evaluated top-3 methods 
under varying gallery sizes in the range from 100 to 4,000 (the whole test {gallery} set).
We observed in Fig. \ref{figure:gallery_size} that
all methods degrade the performance given larger gallery search pools.
When increasing the gallery size
from 100 to 4,000, the mAP performance of NPSM drops from
77.9\% to 53.0\%, i.e. -24.9\% degradation (no reported Rank-1 results).
In comparison, the CLSA is more robust against the gallery size,
with mAP/Rank-1 drop at -9.7\% (77.5-87.2) and -9.1\% (79.4-88.5).
This is primarily because more distracting people are involved in the identity matching process,
presenting more challenging tasks.
Importantly, the performance gain of CLSA over other competitors
becomes even higher at larger search scales,
desirable in real-world applications.
This indicates the superior deployment scalability and robustness
of CLSA over existing methods in tackling a large scale person search problem, 
further showing the importance of solving the previously ignored multi-scale matching challenge
given auto-detected noisy bounding boxes in person search.
%, which has been mostly
%ignored in the literature so far. 

\begin{table}
	%	\scriptsize
	\centering
	\setlength{\tabcolsep}{0.5cm}
	\caption{Evaluation on PRW. The best and second-best results are in red and blue.}
	\label{table:prw}
	\begin{tabular}{|c||c|c|}
		\hline
		Method &Rank-1 (\%) & mAP (\%) \\
		\hline \hline
		ACF-Alex\cite{dollar2014fast}+LOMO\cite{liao2015person}+XQDA\cite{liao2015person} &30.6&10.3\\ 
		ACF-Alex\cite{dollar2014fast}+IDE$_{det}$\cite{zheng2016person} &43.6&17.5\\
		ACF-Alex\cite{dollar2014fast}+IDE$_{det}$\cite{zheng2016person} +CWS \cite{zheng2016person}& 45.2&17.8  \\
		\hline
		DPM-Alex\cite{felzenszwalb2010object}+LOMO\cite{liao2015person}+XQDA\cite{liao2015person}&34.1&13.0 \\
		DPM-Alex\cite{felzenszwalb2010object}+IDE$_{det}$\cite{zheng2016person} &47.4&20.3 \\		
		DPM-Alex\cite{felzenszwalb2010object}+IDE$_{det}$\cite{zheng2016person}+CWS\cite{zheng2016person} &48.3&20.5 \\
		\hline
		LDCF\cite{nam2014local}+LOMO\cite{liao2015person}+XQDA\cite{liao2015person} & 31.1&11.0\\
		LDCF\cite{nam2014local}+IDE$_{det}$\cite{zheng2016person}&44.6&18.3\\
		LDCF\cite{nam2014local}+IDE$_{det}$\cite{zheng2016person} +CWS\cite{zheng2016person}&45.5&18.3 \\			
		OIM\cite{xiao2017joint}&49.9&21.3  \\
		NPSM\cite{liu2017neural}&\bf\color{blue}53.1&\bf\color{blue}24.2\\
		\hline
		\bf CLSA &\bf\color{red}{65.0}&\bf\color{red}38.7 \\
		\hline
	\end{tabular}	
\end{table}

\noindent{\bf Evaluation on PRW }
We further evaluated the CLSA against 11 existing competitors on the
PRW dataset under the benchmarking setting with 11,816 gallery scene images.
Overall, we observed similar performance comparisons with the state-of-the-art methods as on CUHK-SYSU.
In particular,
the CLSA is still the best person search performer
with significant accuracy margins over other alternative methods,
surpassing the second-best model NPSM by
11.9\% (65.0-53.1) and 14.5\% (38.7-24.2) in Rank-1 and mAP, respectively.
%
%Other methods are outperformed by CLSA with larger performance margins.
%
This consistently suggests the model design advantages of CLSA 
over existing person search methods in a different video surveillance scenario.

\begin{table}
	\centering
	\setlength{\tabcolsep}{0.4cm}
	\caption{Evaluating different multi-scale deep learning methods 
		on CUHK-SYSU in the standard 100 sized gallery setting.
		FLOPs: FLoating point OPerations.}
	\label{table:scale_method}
	\begin{tabular}{|c||c|c||c|}
		\hline %\noalign{\smallskip}
		Method &Rank-1 (\%) & mAP (\%) & FLOPs ($\times$10$^9$) \\
		\hline \hline 
		ResNet-50 & 82.5 & 81.6 &\bf 2.678 \\
		
		In-Network Pyramid &81.1&80.2 &\bf 2.678 \\
		\hline 
		%		\color{red}
		DeepMu \cite{qian2017multi} &78.3&75.8 & - \\
		MST \cite{he2014spatial}&82.7&81.9 & 8.034\\
		%Feature fusion&81.5&94.8&96.7&98.2&98.9&80.5\\
		DPFL \cite{chen2017person}&84.7&83.8 & 5.400\\
		FPN \cite{lin2017feature}&85.5&85.0 & 4.519\\
		\hline
		\bf CLSA &{\bf88.5}&{\bf87.2} & 2.680 \\
		\hline 
	\end{tabular}
	%			\vspace{0.9cm}
\end{table}

\subsection{Comparisons to Alternative Multi-Scale Learning Methods}

Apart from existing person search methods,
we further evaluated the effectiveness of CLSA 
by comparing with the in-network feature pyramid (baseline)
and four state-of-the-art multi-scale deep learning approaches
including 
DeepMu \cite{qian2017multi},
MST \cite{he2014spatial}, 
DPFL \cite{chen2017person},
and FPN \cite{lin2017feature}
on the CUHK-SYSU benchmark.
We used the standard 100 sized {gallery} setting in this test.
For all compared methods,
we utilised the same person detection model
and the same backbone identity matching network (except DeepMu that exploits a specially proposed CNN architecture) as the CLSA
for fair comparison.

Table \ref{table:scale_method} shows that the proposed CLSA is more effective 
than other multi-scale learning algorithms
in person search.
In particular, we have these observations:
{\bf (1)} The in-network feature pyramid decreases
the overall performance as compared to 
using the standard ResNet-50 features (no pyramid)
by a margin of 1.4\% (82.5-81.1) in Rank-1 and 1.4\% (81.6-80.2) in mAP.
This verifies our hypothesis that
directly applying the CNN feature hierarchy may
harm the model performance due to the intrinsic
semantic discrepancy across different pyramid levels.
{\bf (2)}
CLSA improves the baseline 
in-network feature pyramid by 
a gain of 
7.4\% (88.5-81.1) in Rank-1 and 
7.0\% (87.2-80.2) in mAP.
This indicates the exact effectiveness 
of the proposed cross-level semantic alignment mechanism
in enhancing the person identity matching capability 
of the CNN feature representation
in an end-to-end learning manner.
{\bf (3)}
Three ResNet-50 based competitors all bring about person search 
performance improvement although less significant than the CLSA.
This collectively suggests the importance of addressing 
the multi-scale matching problem in person search.
{\bf (4)}
For model computational efficiency in FLOPs (FLoating point OPerations) per bounding box,
CLSA has the least (a marginal) cost increase
compared to other state-of-the-art multi-scale learning methods.
This shows the superior cost-effectiveness of CLSA
over alternative methods in addition to its accuracy advantages.

\subsection{Further Analysis and Discussions}

{
	\noindent{\bf Effect of Person Detection }We analysed the effect of person detection
	on the person search performance using the CUHK-SYSU benchmark. 
	We started with the three customised components 
	of Faster-RCNN (Sec \ref{sec:det}).
	Table \ref{table:detection_component} shows that:
	{\bf(1)} The region proposal resizing and max-pool operation does not hurt the model performance.
	In effect, this is a replacement of ROI pooling.
	In the context of an average pooling
	to 1$\times$1 feature map followed, 
	such a design remains the capability of detecting small objects
	therefore imposing no negative effect.
	%	to 1$\times$1 feature map is followed afterwards.
	%
	{\bf(2)} 
	Freezing the first block's parameters in fine-tuning detector helps
	due to the commonality of source and target domain data in
	low-level feature patterns.
	%	the CLSA slightly benefits these specific changes.
	%
	{\bf(3)} 
	Using all sized proposals improves the result. % although not significant
	It is worthy noting this does not reduce the model efficiency, 
	because only top 256 boxes per image are remained after 
	the Non-Maximum Suppression operation, similar to 
	the conventional case of selecting larger proposals.
	There are an average of 6.04 bounding boxes per image on CUHK-SYSU.
}

\begin{table}
	\centering
	\scriptsize%\footnotesize
	\setlength{\tabcolsep}{0.12cm}
	\caption{Detection model component analysis on CUHK-SYSU.}
	\label{table:detection_component}
	\begin{tabular}{|c|c|c||c|c||c|c||c|c|}
		\hline %\noalign{\smallskip}
		& \multicolumn{2}{c||}{Full}
		& \multicolumn{2}{c||}{{No} resize\&max-pool}
		& \multicolumn{2}{c||}{{Not} fix $1^\text{st}$ block}
		& \multicolumn{2}{c|}{{Not} all sized proposals} 
		\\ \hline 
		Metric (\%) 
		&Rank-1 & mAP  
		&Rank-1 & mAP 
		&Rank-1 & mAP  
		&Rank-1 & mAP  \\
		\hline \hline 
		CLSA
		&\bf88.5&87.2
		&88.3&\bf87.3 
		&87.7&86.8
		&87.9&86.9
		\\
		\hline 
	\end{tabular}
\end{table}

\begin{figure}
	\centering
	\includegraphics[width=1.0\linewidth]{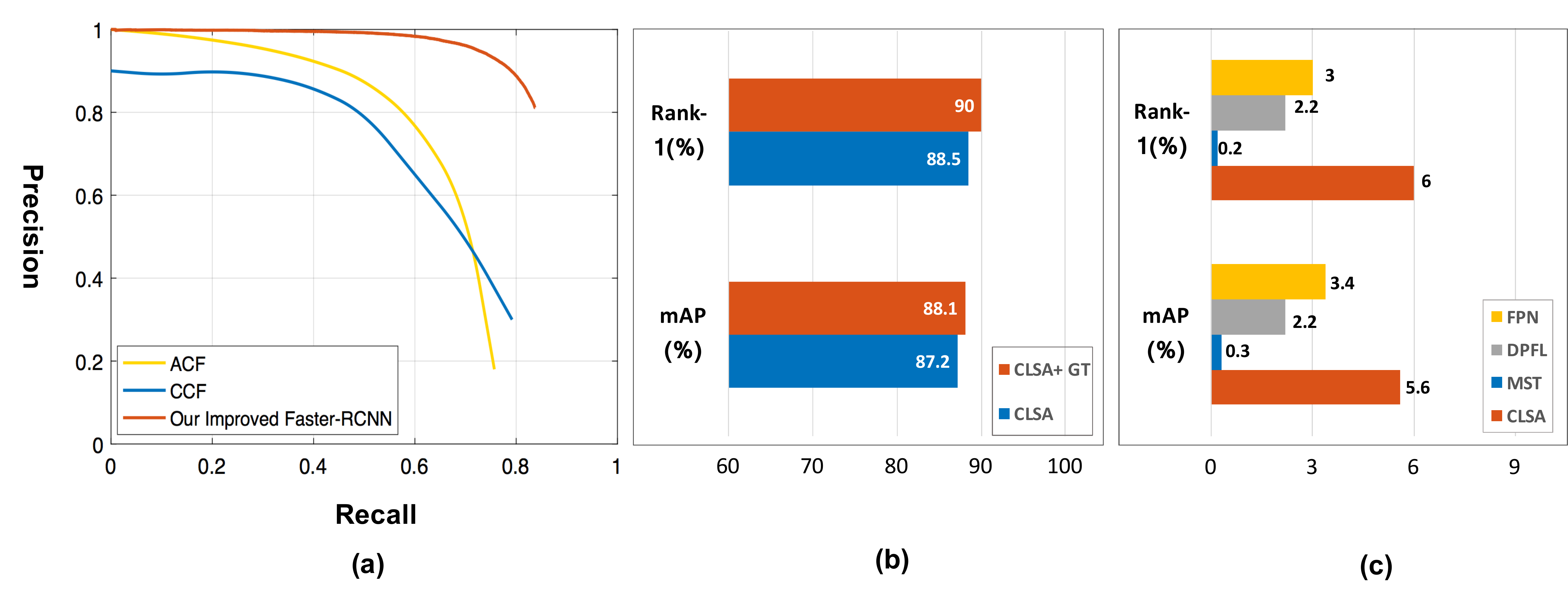}
	\caption{Evaluation of person detection on CUHK-SYSU 
		in the standard 100 sized gallery setting.
		{\bf(a)} Person detection precision-recall performance.
		{\bf(b)} The person search performance of the CLSA
		based on auto-detected {\em or} ground-truth person bounding box images.
		{\bf(c)} Person detection {\em versus} multi-scale learning
		on the effect of person search performance.
	}
	\label{fig:mAP-decreased}	
\end{figure}

We then evaluated the holistic person detection performance
with comparison to other two detection models (ACF \cite{dollar2014fast} and  
CCF \cite{yang2015convolutional}).
%and its person search performance effect.
%by %comparing with using the ground-truth bounding boxes
%in comparison of the CLSA multi-scale learning.
%
%
For person detection, 
it is shown in Fig. \ref{fig:mAP-decreased} (a)
that the precision performance of both ACF and CCF 
drops quickly when increasing the recall rate,
whilst our improved Faster-RCNN 
remains more stable.
This shows the effectiveness of deep learning detectors
along additional model improvement from our CLSA.
This is consistent with the results
in Table \ref{table:cuhk_eval} and Table \ref{table:prw}
that the CLSA outperforms ACF or CCF based methods 
by 20+\% in both rank-1 and mAP.

We further tested the person search effect of our detection model
by comparing with the results based on ground-truth bounding boxes.
It is found in Fig. \ref{fig:mAP-decreased} (b) 
with perfect person detection, the CLSA gives only a gain of 
{0.9}\% (88.1-87.2) in {mAP} and 
1.5\% (90.0-88.5) in {Rank-1}.
This indicates that the person detection component is not necessarily
a major performance bottleneck in person search,
thanks to modern object detection models.
On the other hand, Table \ref{table:scale_method}
also shows that addressing the multi-scale challenge 
is more critical for the overall model performance on person search, 
e.g. CLSA brings a performance boost of 6.0\% (88.5-82.5\%) 
in Rank-1
and 5.6\% (87.2-81.6)
in mAP 
over the baseline network ResNet-50.

%\subsubsection{Which layer need Semantic Alignment} 
\noindent{\bf Effect of Feature Pyramid } 
We evaluated the performance effect of feature pyramid of CLSA on
CUHK-SYSU. %under the standard setting.
Recall that the in-network feature pyramid construction
is based on the selection of ResNet blocks 
(see Sec. \ref{sec:CLSA} and Fig. \ref{fig:scale-reid}).
We tested three block selection schemes:
5-4,
5-4-3 (used in the final CLSA solution),
and 5-4-3-2.
Table \ref{table:semantic choice}
shows that
a three-level pyramid is the optimal.
It also suggests that performing semantic alignment
directly with elementary features such as those
extracted from the Res2X block
may degrade the overall representation
benefit in the pyramid,
due to the hard-to-bridge semantic gap.

\begin{table}
	\centering
	\setlength{\tabcolsep}{0.6cm}
	\caption{Effect of in-network feature pyramid construction on CUHK-SYSU. 
		%		in the standard 100 sized gallery setting.
	}
	\label{table:semantic choice}
	\begin{tabular}{|c||c|c|c|}
		\hline %\noalign{\smallskip}
		Blocks Selection & 5-4 & 5-4-3 & 5-4-3-2 \\
		\hline \hline
		Rank-1 (\%)
		& 87.3 & \bf 88.5& 85.3\\
		\hline
		mAP (\%)
		& 86.2&\bf 87.2 & 84.3\\
		\hline
	\end{tabular}	
\end{table}

\noindent{\bf Effect of Temperature Softness }
We evaluated the impact of the temperature parameter setting in Eq. \eqref{eq:softmax_soft} 
in the range from 1 to 7. 
Table \ref{table:Sensitive of T} shows that this parameter 
is not sensitive with the best value as 3.
%This show semantic alignment is robust for parameters setting.

\begin{table}
	\centering
	\setlength{\tabcolsep}{0.6cm}
	\caption{
		Effect of temperature softness (Eq. \eqref{eq:softmax_soft}) on CUHK-SYSU.
		%		in the standard 100 sized gallery setting.
	}
	\label{table:temperature}
	\label{table:Sensitive of T}
	\begin{tabular}{|c||c|c|c|c|}
		\hline %\noalign{\smallskip}
		Temperature $T$ &1&3&5&7 \\
		\hline \hline
		Rank-1 (\%)
		&88.3&\bf{88.5}&88.3&88.1\\ \hline
		mAP (\%)
		&87.0&{87.2}&\bf87.3&86.9\\
		\hline
	\end{tabular}
\end{table}

{%\color{red}
	\noindent{\bf Evaluating Person Re-ID and Object Classification}
	We evaluated the effect of CLSA on person re-id 
	(Market1501 \cite{zheng2015scalable},
	CUHK03 \cite{li2014deepreid}) 
	and object image classification (CIFAR100 \cite{krizhevsky2009learning}),
	in comparison to ResNet-50.
	Table \ref{table:OtherTask} shows the positive performance gains 
	of our CLSA method on both tasks.
	For example,
	the CLSA improves person re-id
	by $3.5\%$(88.9-85.4) in Rank-1 and $4.5\%$ (73.1-68.6) in mAP
	on Market-1501.
	This gain is smaller than that on the same source video based PRW (see Table \ref{table:prw}),
	due to the potential reason that person bounding boxes of Market-1501 
	have been manually processed with limited and artificial scale variations. 
	Moreover, our method also benefits the CIFAR object classification 
	with a $1.5\%$ (76.2-74.7) top-1 rate gain.
	%	of improvement for R1, mAP respectively
	%	For Person Re-identification, although the performance improvement on Market1501 and CUHK03 is less than person search task due to less scale variation in re-id task (Fig. \ref{fig:scale_change}), it is still remarkable with $3.5\%$  and $4.5\%$ of improvement for R1, mAP respectively.
	These observations suggest the consistent and problem-general advantages of our model
	in addition to person search in unconstrained scene images. 
	%	For Person Re-identification, although the performance improvement on Market1501 and CUHK03 is less than person search task due to less scale variation in re-id task (Fig. \ref{fig:scale_change}), it is still remarkable with $3.5\%$  and $4.5\%$ of improvement for R1, mAP respectively.
}

\begin{table}
	\centering
	%	\scriptsize
	\setlength{\tabcolsep}{0.15cm}
	\caption{Evaluating the CLSA on re-id and object classification benchmarks.}
	\label{table:OtherTask}
	\begin{tabular}{|c||c|c||c|c||c||c|}
		\hline %\noalign{\smallskip}
		Dataset 
		& \multicolumn{2}{c||}{Market-1501 \cite{zheng2015scalable}}
		& \multicolumn{2}{c||}{CUHK03 \cite{li2014deepreid}} 
		& Dataset 
		& CIFAR100 \cite{krizhevsky2009learning}  \\ \hline 
		Metric (\%) &Rank-1  & mAP  &Rank-1  & mAP  & Metric (\%) &Top-1 rate \\
		\hline \hline 
		ResNet-50 & 85.4& 68.6&48.8&47.5&ResNet-110&74.7 \\ \hline
		CLSA  &\color{black}\bf88.9&\color{black}\bf73.1&\color{black}\bf52.3&\color{black}\bf50.9&CLSA &\color{black}\bf76.2 \\
		\hline 	
	\end{tabular}
	%			\vspace{0.9cm}
\end{table}

\section{Conclusion}
In this work, we present a novel {\em Cross-Level Semantic Alignment} (CLSA) 
deep learning framework for
person search in unconstrained scene images. 
In contrast to existing person search methods
that focus on improving the people detection performance,
our experiments show that
solving the multi-scale matching challenge
is instead more significant for improving the person search results. 
To solve this under-studied cross-scale person search challenge,
we propose an end-to-end CLSA deep learning method
by constructing an in-network feature pyramid structural representation 
and enhancing its representational power
with a semantic alignment learning loss function. 
This is designed specially to make all
feature pyramidal levels identity discriminative
therefore leading to a more effective hierarchical representation
for matching person images with large and unconstrained scale variations.
Extensive comparative evaluations have been conducted on two 
large person search benchmarking datasets CUHK-SYSU and PRW. % at varying deployment scales.
The results validate the performance superiority and advantages of
the proposed CLSA model over a variety of state-of-the-art
person search, person re-id and multi-scale learning methods.
%by significant margins. 
%
We also provide comprehensive in-depth CLSA component evaluation and analysis
to give the insights on model performance gain and design considerations.
In addition, we further validate the more general performance advantages of the CLSA method 
on the person re-identification and object categorisation tasks.

\section*{Acknowledgements}
{\small This work was partly supported by the China Scholarship Council, Vision Semantics Limited, the Royal Society Newton Advanced Fellowship Programme (NA150459), and Innovate UK Industrial Challenge Project on Developing and Commercialising Intelligent Video Analytics Solutions for Public Safety (98111-571149).}

\clearpage

\bibliographystyle{splncs04}
\bibliography{person_search_bib}
\end{document}